\documentclass[11pt,a4paper]{article}
\usepackage[hyperref]{acl2019}
\usepackage{times}
\usepackage{latexsym}
\usepackage{amsfonts} 
\usepackage{amsmath} 
\usepackage{amssymb} 
\DeclareMathAlphabet{\pazocal}{OMS}{zplm}{m}{n} 
\usepackage[utf8]{inputenc}
\usepackage{fourier} 
\usepackage{array}
\usepackage{makecell}
\usepackage{graphicx}
\usepackage{algorithm,algorithmic,multicol}

\usepackage{url}

\aclfinalcopy 
\def\aclpaperid{***} 

\title{MML: Maximal Multiverse Learning\\ for Robust Fine-Tuning of Language Models}

\author{Itzik Malkiel \\
  Tel Aviv University \\
  Microsoft \\
\\\And
  Lior Wolf \\
  Tel Aviv University \\
  Facebook AI Research \\
}

\date{}

\begin{document}
\maketitle
\begin{abstract}
Recent state-of-the-art language models utilize a two-phase training procedure comprised of (i) unsupervised pre-training on unlabeled text, and (ii) fine-tuning for a specific supervised task. More recently, many studies have been focused on trying to improve these models by enhancing the pre-training phase, either via better choice of hyperparameters or by leveraging an improved formulation. However, the pre-training phase is computationally expensive and often done on private datasets. In this work, we present a method that leverages BERT's fine-tuning phase to its fullest, by applying an extensive number of parallel classifier heads, which are enforced to be orthogonal, while adaptively eliminating the weaker heads during training. Our method allows the model to converge to an optimal number of parallel classifiers, depending on the given dataset at hand. 

We conduct an extensive inter- and intra-dataset evaluations, showing that our method improves the robustness of BERT, sometimes leading to a +9\% gain in accuracy. These results highlight the importance of a proper fine-tuning procedure, especially for relatively smaller-sized datasets. Our code is attached as supplementary and our models will be made completely public.
\end{abstract}

\section{Introduction}

Recently, there has been an increasing number of studies suggesting the use of general language models, for improving natural language processing tasks \cite{dai2015semi, peters2018deep, radford2018improving, howard2018universal}. Among the most promising techniques, the unsupervised pretraining approach \cite{dai2015semi, radford2018improving} has emerged as a very successful method, that achieves state-of-the-art results on many language tasks, including sentiment analysis \cite{socher2013recursive}, natural language inference \cite{williams2017broad} and similarity and paraphrase tasks \cite{dolan2005automatically, cer2017semeval}. This approach incorporates a two-phase training procedure. The first phase utilizes an unsupervised training of a general language model on a large corpus. The second phase applies supervision to fine-tune the model for a given task.

More lately, unsupervised pretraining models such as BERT \cite{devlin2018bert}, XLNET \cite{yang2019xlnet} and RoBERTa \cite{liu2019roberta}, have achieved unprecedented performance, even exceeding human level of performance on some language tasks. For example, in the GLUE benchmark \cite{wang2018glue}, BERT \cite{devlin2018bert} reported to achieve performance that exceeds human level on a few different datasets, such as QNLI \cite{rajpurkar2016squad}, QQP \cite{chen2018quora} and MRPC \cite{dolan2005automatically}. However, although the great progress achieved by these task-specific and dataset-specific models, it is not yet clear how well they can generalize to different tasks, and how robust they are when evaluating the same task on different datasets. 

The most direct way to estimate the specificity of a learned model is by employing cross-benchmark experiments. These evaluations can be done by using datasets of the same task the model was specialized on (to measure robustness), or by utilizing datasets from different tasks (to measure generalization).

In our work, we build upon the multiverse method of \cite{littwin2016multiverse}, which was shown to lead to cross dataset robustness in the computer vision task of face recognition as well as on the CIFAR-100 small image recognition dataset. The multiverse loss generalizes the cross entropy loss, by simultaneously training multiple linear classifiers heads to perform the same task. In order to prevent multiple copies of the same classifier, in the multiverse scheme, each classifier is mutually orthogonal to the rest of classifiers. The number of multiverse heads used was limited, never more than seven and typically set to five.

We propose a novel fine-tuning procedure for enhancing the robustness  of recent unsupervised pretraining language models, by employing a large number of multiverse heads. The essence of our technique is as follows: given a pretrained language model and a downstream task with labeled data, we fine-tune the model using a maximal number of multiverse classifiers. The fine-tuning goal is to both minimize the task loss and an orthogonality loss applied to the classifier heads. When enforcing orthogonality hinders the classifiers' performance, we detect and eliminat the less effective classifier heads. 

The technique therefore preserves a maximal set of classifiers, which comprises of the best performing ones. By maintaining this maximal subset during training, our method leverages multiverse loss to its fullest. Hence, we name our method Maximal Multiverse Learning (MML).

Our contributions are as follows: (1) we present MML, a general training procedure to improve the robustness of neural models. (2) We apply MML on BERT and report its performance on various datasets. (3) we propose a set of cross dataset evaluations using common NLP benchmarks, demonstrating the effectiveness of MML in comparison to regular BERT fine-tuning. 

\section{Related Work}

Recent breakthroughs in the field of NLP are centered around unsupervised pretraining of language models. The different variants  can be categorized by two main approaches: (1) feature-based models, such as \cite{peters2018deep} and (2) fine-tuning models, such as \cite{devlin2018bert, liu2019roberta,yang2019xlnet}. The former technique utilize a language neural based model as a feature extractor. The extracted features may be used for the training of another separate models, receiving the extracted features as input. The second approach, utilizes a similar pre-trained model, but fine-tune it in an end-to-end manner to specialize on a given task. During the fine-tuning phase, all of the parameters of the model are updated, as a relatively small number of parameters are trained from scratch.

The usage of multiple classifiers can be found in few places in the literature. In GoogLeNet \cite{43022}, the authors use multiple classifier heads in different places in the model architecture. The additional classifiers led for better propagation of gradients during training. However, with the advent of better conditioning and normalization methods, as well as with the modern introduction of skip connections in architectures such as the ResNet \cite{he2016deep}, the practice of adding intermediate branches for the sake of introducing loss at lower levels was mostly abandoned.

The multiverse loss was shown to promote better transfer learning and to lead to a low-dimensional representation in the penultimate layer~\cite{littwin2016multiverse}. However, the current literature does not present any methodological way to select the number of multiverse heads and the idea was only applied for a handful of parallel classifiers. 

In MML, hundreds of multivese heads are used. An emphasis is put on the resulting multi-term loss settings, in which the classifier accuracy is contrasted with the orthogonality constraint. MML balances the two terms by pruning the multiverse classifiers that underperform during training.

\section{Method}
The section presents the problem setup, the MML architecture and loss terms, and training algorithm.

\begin{figure*}[t]
\includegraphics[width=1.0\linewidth]{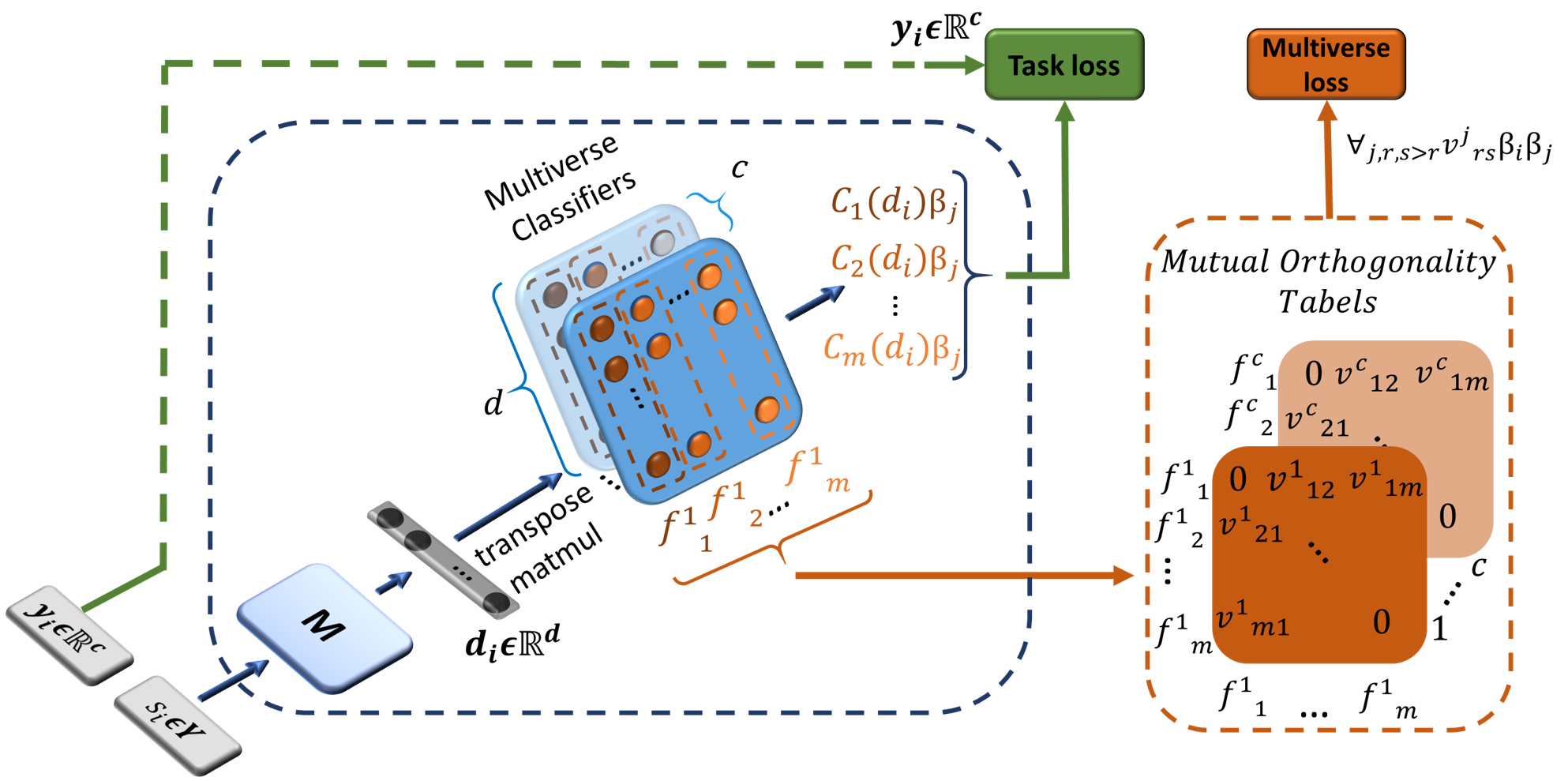}
\centering
\caption{A schematic illustration of the MML model. The task loss comprises a loss-term for each multiverse classifier, using the given labels of the task at hand. The mutual orthogonality tables hold the absolute value of the dot product calculated between the weights of all classifiers, across the different classes. Since orthogonality of a classifier with itself is ignored, we set the diagonal to 0. Following multiverse loss definition and since orthogonality is symmetric, only half of each table values are passed to the multiverse loss. }
\label{fig:MML-model}
\end{figure*}

\subsection{Problem Setup}
Let $\mathcal{W}=\left\{w_{i}\right\}_{i=1}^{w}$ be the vocabulary of all supported tokens in a given language. Let $Y$ be the set of all possible sentences that can be generated by $\mathcal{W}$. $Y$ may also contain the empty sentence. We will define $M: Y \times Y \rightarrow \mathbb{R}^{d}$ to be a language model, receiving pairs of elements from $Y$ and returning coding vectors of $d$ dimensions. Given a dataset with $n$ training samples, $s_1...s_n \in Y \times Y$, each associated with a label $y_i \in [1...c]$, we will denote the coding vector of each sample by $d_i := M(s_i)$ where $d_i \in \mathbb{R}^{d}$. As a concrete example, for the BERT model, $d_i$ is the latent embedding of the CLS token. 

Common language models use classifier $C: \mathbb{R}^{d} \rightarrow \mathbb{R}^{c}$ which projects the coding vectors $d_i \in \mathbb{R}^{d}$ by a $d \times c$ matrix, $F_{d \times c}=[f_1, ..., f_c]$, ($f_i \in \mathbb{R}^{d}$), and then adds a bias term $b \in \mathbb{R}^{c}$:
\begin{equation}
\label{single_head}
C(d_i) = d_{i}^{T} F_{d \times c} + b
\end{equation}
The output of $C$ is a logit vector, that will be used to produce probabilities via a softmax function, that can also be expressed as: 
\begin{equation}
p_{i}\left(y_{i}\right)=\frac{e^{d_{i}^{\top} f_{y_{i}}+b_{y_{i}}}}{\sum_{j=1}^{c} e^{d_{i}^{\top} f_{j}+b_{j}}} = \frac{e^{C(d_i)_{y_{i}}}}{\sum_{j=1}^{c} e^{C(d_i)_{j}}}
\end{equation}

Different from other language models that uses a single classifier $C$ as defined above, our model utilizes a multiverse classifier $\mathcal{C}: \mathbb{R}^{d} \rightarrow \mathbb{R}^{c \times m}$ defined as: 
\begin{equation}
\mathcal{C}(d_i) = \left( C_1(d_i), ..., C_m(d_i)\right)
\end{equation} 
where $m$ is a multiplicity parameter,  $\left\{C_j:\mathbb{R}^{d} \rightarrow \mathbb{R}^{c}\right\}_{j=1}^{m} $ are parallel classifiers, each with different weights, applying the same function as Eq.~\ref{single_head}. :
\begin{equation}
C_j(d_i) = d_{i}^{T} F_{d \times c}^j + b_j
\end{equation}

Additionally, we will define $B=\left\{\beta_{j}\in \{0,1\}\right\}_{j=1}^{m}$ as a set of binary scalars. Each classifier head $C_j$ will be associated with a different binary scalar $\beta_j$. The binary scalars will be set with concrete values during training (see Sec.\ref{training}). In our experiments we set m to be equal to the coding vector size $d$, which entails a full rank of active multiverse classifiers at the beginning of the training. All in all, our aggregated model composed of ($M$, $\mathcal{C}, B$).

\begin{algorithm*}[h] 
\label{trainAlgo}
  \caption{Maximal Multiverse Training. Parameters: $K = 1000$, $\gamma = 0.99$, $Threshold = 5$, $\alpha = 2 \cdot e^{-5}$}
    \begin{algorithmic}
      \STATE  $\beta_{j} \gets 1$, $a_{j} \gets 0$,  $\forall 1 \leq j \leq m$
      \FOR{$step$ $\leq$ training\_steps} {
        \STATE Sample a minibatch \{($s_i, y_i$)\}$_{i=1}^{t}$ 
        
        \STATE $MC_\theta$ $\gets$  $  \nabla _\theta$  $\left[
        \lambda \cdot \pazocal{L}_{mv}\left( \mathcal{C}, B\right)
        +
        \frac{1}{t}\sum_{i=1}^{t} \pazocal{L}_{task}\left(M,\mathcal{C},B,\{(s_i,y_i)\}_{i=1}^{t}\right) 
        \right]$ 
        
        \FOR{ $1 \leq j \leq m $} {
        
            \STATE $a_j \gets (1-\gamma) \cdot a_j + \left(\gamma \cdot \frac{1}{t}\sum_{i=1}^{t} \pazocal{L}_{task}\left(M,\mathcal{C},B,\{(s_i,y_i)\}_{i=1}^{t}\right)\right)$
        } \ENDFOR
      
        \STATE $\theta$ $\gets$ $\theta$ +  Adam($\theta$, $MC_\theta$, $\alpha$) 
        \IF{$step \% K == 0$ and $Threshold \leq \sum_{j=1}^{m} \beta_j $}{
            \STATE clusters $\gets$ MeanShift$\left(\left\{a_j | \beta_j =1 \right\}_{j=1}^m\right)$

            \IF{ $|clusters| \geq 2$}{
                \FOR{ $1 \leq j \leq m $} {
                    \STATE $\beta_j \gets 0$, if $a_j \notin min(clusters)$ \; 
                    }\ENDFOR

       }\ENDIF       }\ENDIF
    } \ENDFOR
    \end{algorithmic}
\end{algorithm*}

\subsection{The Loss Function}
Our loss function is composed of two components, the task loss and the multiverse loss. The task loss is set according to the task in hand, and its essence is optimizing the performance of all active multiverse classifiers, each independently, using the supervision obtained by the given labels. The active multiverse classifiers are the ones that survive the dynamic elimination used during training (see Sec.~\ref{training} for more details), and are associated with a value $\beta_j=1$. The multiverse loss soft-enforces orthogonality among the active classifiers, and its purpose is to regularize the model by encouraging $M$ to produce coding vectors that are robust enough to be effective for a large number of orthogonal classifiers.

As mentioned earlier, each classifier $C_j$ is associated with a binary value $\beta_j$, which controls the applicability the classifier and is configured during training. Under the context of the loss function, setting $\beta_j$ to $0$ would eliminate the impact of the $j^{th}$ classifier head $C_j$ for both the task loss and multiverse loss.

For a multi-class classification task we apply the following task loss: 

\begin{equation}
\label{main_loss}
\pazocal{L}_{task} = -\Sigma_{i=1}^{n}\Sigma_{j=1}^{m} \pazocal{L}^j_{cce}   \beta_j
\end{equation}
where n is the number of training samples, and $\pazocal{L}^j_{cce}$ is the cross entropy loss 
\begin{equation}
\pazocal{L}^j_{cce} = y_i log(C_j(M(s_i))_{y_{i}})
\end{equation}
For a binary classification task we set  $\mathcal{C}: \mathbb{R}^{d} \rightarrow \mathbb{R}^{2 \times m}$ and use the same loss from Eq.~\ref{main_loss}. For a regression task, we replace $\pazocal{L}^j_{cce}$ with $\pazocal{L}^j_{L2}$: 
\begin{equation}
\pazocal{L}^j_{L2} = \left\|y_i - C_j(M(s_i))\right\|_{2}^{2}  
\end{equation}

The second loss term enforces orthogonality between the set of classifiers, for each class separately. In our work, orthogonality is being forced through the weights of each classifier, using the multiverse loss: 
\begin{equation}
\pazocal{L}_{mv} =  \Sigma_{j,r, s>r}  \left| f_{j}^{r \top} f_{j}^{s}  \beta_r   \beta_s \right|
\end{equation}
where $f_j^r$ is the $r$th column of the weights matrix corresponds to classifier $C_j$. As motioned, we use $\beta_j$ in order to allow the training algorithm to dynamically eliminate the less effective multiverse classifiers during training (see Sec.~\ref{training}). 

The total loss $\pazocal{L}_{total}$ is defined as:
\begin{equation}
\pazocal{L}_{total} = \pazocal{L}_{task} + \lambda \pazocal{L}_{mv}
\end{equation}

Similar to \cite{littwin2016multiverse}, we set $\lambda = 0.005$, throughout all of our experiments. The MML model is illustrated in fig \ref{fig:MML-model}.

\subsection{Maximal Multiverse Training}
\label{training}

The training algorithm begins with an initialization of the aggregated model ($M, \mathcal{C}, B$). $M$ may be initialized by any pre-trained general language model. The multiverse classifiers are randomly initialized from scratch, and all classifiers are initially activated by setting $\beta_j = 1$ for $ \forall \beta_j \in B $.

During training, we track the performance of each multiverse classifier separately. Every $K$ steps, we search for a subset of the top-performing classifiers. When we find such a subset, we eliminate the less performing classifiers by setting their corresponding $\beta$s to 0.

In order to detect the top-performing subset of classifiers, we calculate a moving average variable $a_j$ for each multiverse classifier. Specifically, $a_j$ holds the moving average of the task loss value $\pazocal{L}^j_{task}$ associated with classification head $C_j$. $a_j$ is being updated for every training step, using the moving average momentum constant of 0.99.

During training and every $K$ steps, we run MeanShift algorithm \cite{comaniciu2002mean} on the set $\left\{a_{j}| \beta_j=1 \right\}$. MeanShift is a clustering algorithm that analyzes the underlying density function of the samples. The algorithm reveals the number of clusters in a given data, and retrieves the corresponding centroid for each detected cluster. By utilizing MeanShift, we define the subset of top-performing multiverse heads as the cluster associated with the minimal centroid value. Next, we eliminate the rest of the multiverse heads by setting their corresponding $\beta$ to 0. This adaptive elimination is stopped when we reach a minimal number of active heads, see Alg.~1.

\subsection{Inference}

At inference, we use the active multiverse heads to retrieve predictions. Specifically, given a sample $s_i$, we calculate the logits $\hat{y}$ as:
\begin{equation}
\hat{y} := \frac{\sum_{j=1}^{m} C_j(M(s_i)) \cdot \beta_j}{\sum_{j=1}^{m} \beta_j}  
\end{equation}
for classification tasks, we apply the softmax function on $\hat{y}$, and return its output. For regression tasks, we simply return $\hat{y}$.

\section{Results}

\begin{table*}[t!]
\centering
\begin{tabular}{l c c c c c c c c c c}
\textbf{Model} &  \textbf{MNLI} &  \textbf{QQP} & \textbf{QNLI} & \textbf{SST-2} & \textbf{CoLA} & \textbf{STS-B} & \textbf{MRPC} & \textbf{RTE} & \textbf{Average} \\
& 392k & 363k & 108k & 67k & 8.5k & 5.7k & 3.5k & 2.5k & - \\
  \hline

\multicolumn{3}{c} {\makecell{\textit{Single-task single models on dev}}}{} \\
  BERT & 86.6/- & 91.3 & 92.3  & 93.2 &  60.6  & 90.0  & 88.0  & 70.4 & 84.0  \\

  MV-5 & 87.0/-  & 91.4  & 92.2 & 94.0 &  64.3 & 91.1 & 88.0  & 75.4  & 85.4 \\
  MV-1024 & 86.2/-  & 90.5  & 92.2 & 93.6 & 57.9 & 90.6 & 89.0 & 80.1 & 85.0 \\
 
  MML & 87.2/-  & 91.7  & 93.0 & 93.8 & 64.5 & 91.1 & 89.0 & 80.1 & 86.3 \\

\hline
\multicolumn{3}{c} \makecell{\textit{Single-task single models on test}} \\

  BERT & 86.7/85.9 & 89.3 & 92.7 & 94.9 & 60.5 & 86.5 & 85.4 & 70.1 & 83.55\\
 
  MML & 87.0/86.0 & 89.4 & 92.6 & 94.6 & 58.6  & 88.1 &  86.7 & 74.2 & 84.13
  \\

  \hline
 
  MML \#heads  & 5  & 14  & 23 & 979 & 31 & 45 & 913 & 1024 & - 

\end{tabular}
\caption{results on GLUE benchmark \cite{wang2018glue}. Bert results taken from \cite{devlin2018bert}. Accuracy scores are reported for all datasets, except STS-B, for which Spearman Correlation is reported. The last row exhibits the number of active multiverse heads of the converged MML model. For example, for MRPC, our MML model used 913 active multiverse heads, while for MNLI it maintained only 5.}
\label{Tab:glue}
\end{table*}

In this study, we evaluate MML, applied on a pre-trained BERT \cite{devlin2018bert} model, using nine NLP datasets while employing two different settings: (1) a straight forward fine-tuning on different downstream tasks from the GLUE benchmark \cite{wang2018glue}, and (2) cross dataset evaluations for different datasets of the same or similar task.

For the first, we fine-tune MML on each dataset separately, and evaluate its performance on the development set and the test set of the same dataset. For the second, we evaluate our fine-tuned MML models on the train and development set of other datasets within the same task category. This allows us to study the robustness level of all models, across different datasets. 

In addition, we perform an ablation study and report empirical results that showcase the efficiency of MML and its variants, compared to a baseline BERT. 

\subsection{The Datasets}
\label{dataset-ctegories}
We adopt 8 datasets from the GLUE benchmark \cite{wang2018glue}, and one extra dataset supporting the task of Natural Language Inference (NLI). The datasets can be arranged by categories as follows.

\subsubsection{Inference Tasks} 
In this category, we utilize three datasets from the GLUE benchmark \cite{wang2018glue}, along with an external dataset named SNLI \cite{bowman2015large} that shares the same task. 

\textbf{RTE} The Recognizing Textual Entailment dataset \cite{bentivogli2009fifth} is composed of sentence pairs gathered from various online news sources. The task is to predict whether the second sentence is an entailment of the first sentence (binary classification).

\textbf{MNLI} Multi-Genre Natural Language Inference Corpus \cite{williams2017broad} is a dataset comprised of sentence pairs with textual entailment annotations. For each pair of sentences, the task is to determine whether the second sentence is a contradiction, neutral or entailment with respect to the first one (multiclass classification).

\textbf{SNLI} The Stanford Natural Language Inference dataset \cite{bowman2015large} also contains sentence pairs. The task here is identical to MNLI, with the same three labels (multiclass classification). However, the two datasets were gathered from different sources.

\textbf{QNLI} The Question-answering Natural Language Inference dataset \cite{rajpurkar2016squad} contains question-sentence pairs. The task is to determine whether a sentence contains the answer to its corresponding question (binary classification). 

\subsubsection{Similarity and Paraphrase Tasks} 
This category contains three datasets.
 
\textbf{MRPC} Microsoft Research Paraphrase Corpus \cite{dolan2005automatically} is a dataset of sentence pairs taken from online news websites. The task is to determine whether a pair of sentences are semantically equivalent (binary classification). 

\textbf{QQP} Quora Question Pairs \cite{chen2018quora} is a dataset of questions pairs taken from Quora website. The goal is to determine whether a pair of questions are semantically equivalent (binary classification).

\textbf{STS-B} Semantic Textual Similarity Benchmark \cite{cer2017semeval} is a dataset composed of sentence pairs extracted from news headlines, video and image captions, and natural language inference data. Each pair is annotated with a score between 1 and 5, indicating the semantic similarity level of both sentences. The task is to predict these scores (regression).

\subsubsection{Misc. datasets} 
There are two datasets in this category. The two datasets are not used for the cross dataset evaluation, due to the lack of commonality between their tasks.

\textbf{CoLA} The Corpus of Linguistic Acceptability dataset \cite{warstadt2018neural} consists of examples of expert English sentence acceptability judgments, which were drawn from multiple books. Each sample in this dataset is a string containing English words annotated by whether it is a grammatically sentence of English (binary classification). 

\textbf{SST-2} The Stanford Sentiment Treebank \cite{socher2013recursive} is a dataset composed of sentences extracted from movie reviews. The sentences are assigned with a human annotations of their sentiment, and the task is to determine whether the sentiment of each sentence is positive or negative (binary classification).

\begin{figure*}[t]
\includegraphics[width=1.0\linewidth]{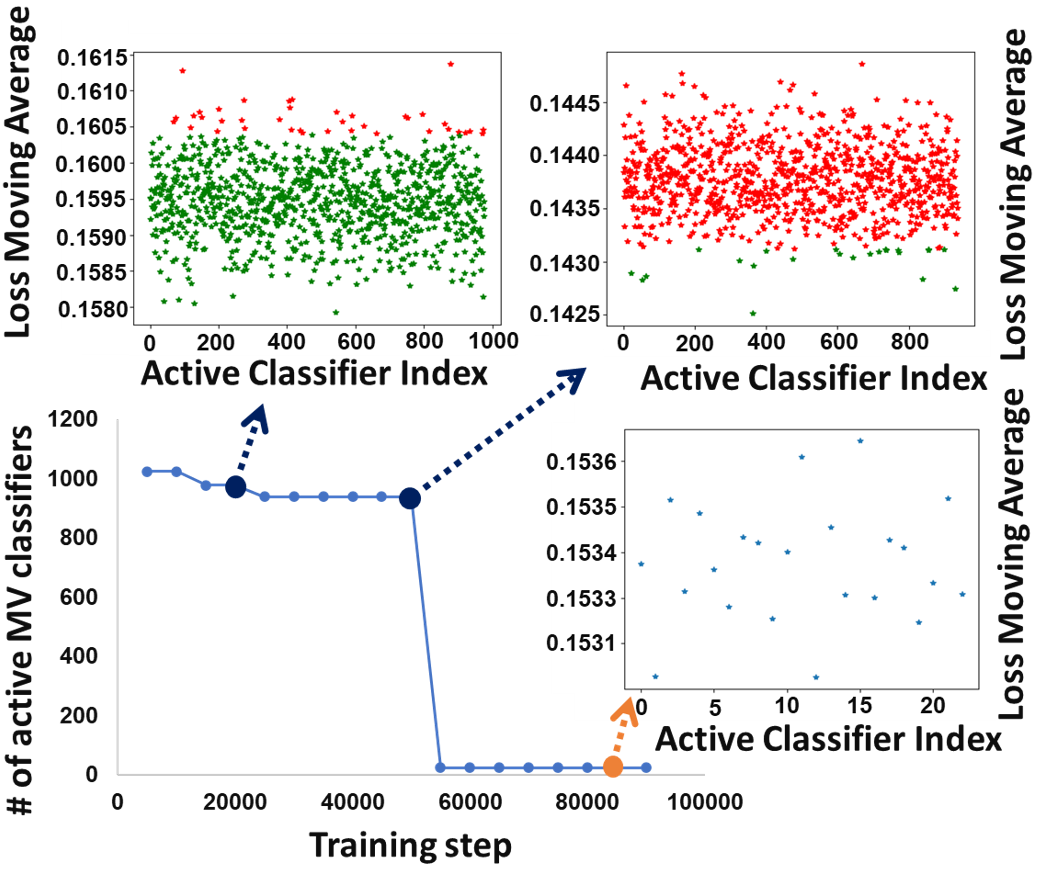}
\centering
\caption{ The number of active multiverse classifiers, per training step, for MML model trained on QNLI. The MeanShift algorithm detects multiple clusters, for three times during training. The two upper plots present the selection of the top-performing heads (green stars), and the elimination of the less performing heads (red stars). The Y values are the moving average calculated on the multiverse heads' loss function. Our MML-QNLI model reaches local minima at step 85K, for which 23 heads were activated. The bottom right plot shows the moving average values of the activated 23 multiverse heads, by the same training step.}
\label{fig:qnli-convergance}
\end{figure*}

\subsection{evaluation on GLUE benchmark}

We evaluated MML on the eight different datasets from the GLUE benchmark, and compared to BERT \cite{devlin2018bert}. In addition, we conduct an ablation analysis for our method, presenting the importance of our Maximal Multiverse Training, which allows the training to adapt the number of multiverse classifiers to each dataset. The ablation disables the classifier elimination step during training, and utilizes the same MML architecture with a fixed number of heads. Each models was trained and evaluated on a single dataset. Development and test set performance are being reported for each model.

\subsubsection{The Models}
The BERT model we are using is the BERT-Large model from \cite{devlin2018bert}. It contains 24 attention layers, each with 16 attention heads with a hidden layer size of $d=1024$ dimensions. The model was pre-trained using sentence pairs, to both reconstruct masked words and to predict whether sentence pairs are consecutive. BERT's fine-tunning for downstream tasks employs supervision obtained by the given labels of each dataset. 

MML utilizes a pre-trained BERT-Large model, and fine-tunes it via Maximal Multiverse Training, to minimize the $\pazocal{L}_{total}$ loss presented above. During training, the MML model is initialized with $m=1024$ active multiverse classifiers, which is equal to the hidden layer size $d$. During training, the model converges to a smaller  number of multiverse classifiers. 
The number of active multiverse classifiers of each model are presented in last row of Tab. \ref{Tab:glue}. 

Tab,~\ref{Tab:glue} presents the results for the following models: (1) BERT (used as a baseline), (2) MML, (3-4) ablation models of two multiverse models utilizing a fixed number of multiverse classifiers, with 5 and 1024 classifiers, respectively.

As can be seen in the table, compared to BERT, MML yields significantly better results on the test set of four out of eight datasets. The largest gains were reported in the relatively smaller sized dataests, such as RTE, MRPC and STS-B, for whom MML yileds an absolute improvement of 4.1, 1.3, 1.6 points, respectively. This can be attributed to the ability of MML to encourage a more robust learning. On the rest of the datasets, MML yields similar performance on the test, besides CoLA for which a degradation of 1.9 points is reported. On the development set, MML outperforms BERT on all datasets, sometimes by a large margin. Specifically, for RTE and CoLA, MML yields an improvement of almost ten and four points, respectively. 

The ablation models MV-5 and MV-1024, utilize a fixed number of multiverse heads during the entire training. We have found that this hyper parameter can be crucial for model convergence, and when not initialized properly, may significantly reduce performance for the given task in hand. Specifically, for the CoLA dataset, MV-1024 and MV-5 yield a relative performance gap of more than 11\%, in favor of MV-5, while in RTE, there is a gap of 6.2\% in favor of MV-1024. When comparing both MV-5 and MV-1024 to MML, MML produce better or similar performance on the development set of all datasets. More specifically, on RTE and MRPC, MML yields similar performance as in MV-1024, and outperforms it on all the other six datasets. Compared to MV-5, MML yields significantly better performance on four datasets out of eight, and produce similar performance on the rest.

Fig.~\ref{fig:qnli-convergance} presents the amount of active multiverse heads when applying MML on QNLI dataset. During the training of MML-QNLI model, the MeanShift algorithm detected multiple clusters at three times\footnote{ The elimination is being invoked every time the MeanShift algorithm detects multiple clusters. Specifically, for MML-QNLI experiment, multiple clusters appeared three times during the training process. }, through the entire training. Each time, the model eliminated the less performing subsets, and kept the top-performing multiverse classifiers as the active set of classifiers. The model achieved best performance on the development set at training step 85K. At this step, MML-QNLI model utilized 23 active multiverse heads. The plots in the figure present the cumulative loss of each multiverse head, sorted through the X axis according to the indices of the active heads. The red stars associated with the classifiers heads that were eliminated, and the green stars are the heads that were selected as the top-performing subset.

\subsection{Cross Dataset Evaluations}

\begin{table*}[t!]
\centering
\begin{tabular}{|l|c|c|c|c|c|}
\hline
\thead{Model} & \thead{RTE}  & \thead{MNLI}  &  \thead{QNLI} &   \thead{SNLI}  & \makecell{\thead{Average\\ Cross Dataset\\ Improvement \\ Obtained \\ by MML} } \\  \hline\hline
BERT-RTE     &  96.06/70.39   & 69.42/69.17  & 52.46/52.84 & 68.02/69.87 & - \\
MML-RTE   &   99.39/\textbf{80.14} &  79.24/78.42 & 50.86/51.30  & 80.85/82.53 & +9.9\%/+9.5\% \\
\hline
BERT-MNLI       &   79.15/76.89  &  99.59/86.58 & 49.88/51.05 & 81.65/83.65  & -\\

MML-MNLI     & 79.35/78.70 & 99.74/\textbf{86.62} & 49.64/50.22 & 82.37/83.94& +0.21\% /+0.35\% \\
\hline
BERT-QNLI   &  53.37/48.73  &  59.76/59.89  & 99.99/\textbf{94.01}  & 59.33/60.03  & - \\
MML-QNLI  & 53.41/53.79  &  64.93/63.85  & 95.75/92.86 & 62.13/63.58  & +4.48\%/+7.63\%\\
\hline
\end{tabular}
\caption{Cross dataset evaluation for Language Inference tasks. Train/development accuracy are reported separately for each dataset. Each model (a row in the table) was trained on a single dataset denoted by its name, and was evaluated on the train/development sets of all four datasets. The last columns indicates the relative average improvement obtained by MML compared to BERT, and averaged across the three hold-out datasets. BERT models were reproduced with the same hyperparamters used for MML (all BERT reproductions result with similar or better performance compared to the original BERT work \cite{devlin2018bert}).}
\label{Tab:crossInf}
\end{table*}

\begin{table*}[t!]
\centering
\begin{tabular}{|l|c|c|c|c|}
\hline
\thead{Model} & \thead{QQP}  &\thead{MRPC}  & \thead{STS-B*}  & \makecell{\thead{Average\\ Cross Dataset\\ Improvement \\ Obtained \\ by MML}} \\
\hline\hline

BERT-QQP    &   99.73/91.57  &  66.90/68.85  &  88.34/90.12 & - \\

MML-QQP &   99.74/91.68   & 67.77/68.87 & 89.11/90.55 & +1.08\%/+0.25\% \\

\hline
BERT-MRPC  &  65.28/65.18  &  99.37/87.25 & 82.53/88.58&  -\\

MML-MRPC  & 68.37/68.15  & 99.23/88.97  &  86.12/91.32 & +3.42\%/+3.78\%\\
\hline

BERT-STS-B* & 73.13/73.11  & 75.59/75.49  & 100.0/95.49  & -\\
MML-STS-B* & 74.13/74.40 & 75.51/77.94  & 99.85/96.70 & +0.63\%/+2.50\% \\
\hline
\end{tabular}
\caption{cross dataset evaluation for similarity and paraphrase tasks. STS-B* is the modified version of STS-B that forms a binary classification dataset (insterad of regression). STS-B* models were trained as binary classifiers, on STS-B* data. Accuracy scores are reported through all evaluations. The last column present the relative cross dataset improvement obtained by MML, compared to BERT. }
\label{Tab:crossSimilarity}
\end{table*}

To study the robustness level of all models, we perform cross dataset evaluations. In these evaluations, we use the fine-tuned MML models from Tab.~\ref{Tab:glue}. For each model trained on a dataset from the two first categories above (Sec.~\ref{dataset-ctegories}), we evaluate the model on all datasets from the same category. 

Train and development set performances are reported to give a clear view on the robustness and stability of the models, and also to exhibit the level of overfitting when evaluating on the same dataset each model was trained on.

In order to conduct a clean comparison, we finetune BERT with the same hyperparmeters used for MML. Specifically, for the MML we employ 10 epochs for the relative larger datasets, 30 epochs for the medium sized datasets, and 100 epochs for the smaller-sized datasets. All models were trained with a batch size of 32, and a learning rate of 2e-5. Our code can be found at \url{https://github.com/ItzikMalkiel/MML}.

\subsubsection{Cross Inference Datasets Evaluation}

First, we present performance on different inference datasets. We fine-tune both BERT and MML on each dataest separately, and evaluate on four NLI datasets: RTE, MNLI, SNLI, QNLI. Since MNLI and SLNI are multicalss classification tasks with 3 classes, we collapse the labels "neutral" and "contradication" into one label ("non entailment"). This modification, applied only in inference, allows us to evaluate MNLI and SNLI models on RTE and QNLI  models, and vice versa.

The results are reported in Tab.~\ref{Tab:crossInf}. As can be seen, MML exhibits a significantly improved robustness compared to BERT. Each row in the table represent MML or BERT model trained on a single dataset associated by its name. All models are evaluated on all four datasets. In the last column, we report the relative average improvement obtained by MML, calculated by the performance ratio between MML and BERT across all three holdout datasets. For example, for RTE, our MML-RTE model yields 9.9\% relative average performance on the train set of MNLI, QNLI and SNLI, and a 9.5\% average improvement on the development set of these datasets.

\subsubsection{Cross Similarity and Paraphrase Datasets Evaluation}

Next, we conduct cross dataset evaluations on the three datasets for the similarity and paraphrase task. We fine-tune MML and BERT for the datasets MRPC, QQP and STS-B. More specifically, to allow cross evaluations between these models, and since STS-B is a regression task benchmark, while MRPC and QQP address a binary classification task, we adapt STS-B to form a binary classification task. The adaptation is being done by collapsing the labels in the range 1-2 (4-5) to the value of 0 (1). In addition, we omit all the ambiguous samples associated with label values between 2 and 4. This modification to STS-B allows us to identify a distinct set of similar and non-similar sentence pairs. The modified STS-B forms a binary classification dataset with ~3.5K samples. 

As can be seen in Tab.~\ref{Tab:crossSimilarity}, MML yields better performance on the cross evaluations for the similarity and paraphrase datasets. Similar to Tab.~\ref{Tab:crossInf}, each row represent a single model trained on a single dataset. We evaluate all models on all three datasets, and report the average relative improvement obtained by MML calculated on the two holdout datasets. We have found MML to produce improved performance for all models, for example, MML-MRPC yields a ~+3.5\% average improvement calculated on both train and development sets across STS-B* and QQP.

\subsection{Discussion of results}

As can be seen in both Tab.~\ref{Tab:crossInf} and \ref{Tab:crossSimilarity}, conducting the cross dataset evaluations reveals a significant gap in performance for all models when evaluated on holdout datasets, although the holdout datasets share the same or similar task each model was trained for. For example, both MML-MRPC and BERT-MRPC models yield a ~20\% degradation in absolute accuracy on RTE dataset.  Yet, compared to BERT, our MML method produces significantly better performance on the cross evaluations. Specifically, when evaluated on QQP, MML-MRPC outperforms BERT-MRPC by a relative improvement of ~4.6\%, for both development and train set.

Perhaps unintuitively, there is no direct link between the improvement obtained on the same dataset evaluation to that obtained in the cross dataset one. For example, our MML-QNLI model was able to outperform BERT-QNLI in the cross dataset evaluation, although it exhibits a somewhat degraded performance on QNLI's development set and test set. We attribute this to the ability of MML to encourage the model to produce more robust coding vectors.

\section{Conclusion}

In this work, we introduce MML: a method for fine-tuning general language models, that is based on Multiverse loss. MML utilizes a large set of parallel multiverse heads, and eliminates the relatively weaker heads during training. The heads eliminations, employed through the entire course of training, assures the use of a maximal set of top-performing multiverse heads. 

We demonstrate the effectiveness of MML on nine common NLP datasets, by applying inter- and intra- datasets evaluation, where it is shown to outperform the originally introduced BERT model. our results shade light on the robustness level of both models, and showcase the ability of MML to yield improved robustness.

\bibliography{acl2019}
\bibliographystyle{acl_natbib}

\end{document}